\documentclass[10pt,twocolumn,letterpaper]{article}

\usepackage{cvpr}
\usepackage{times}
\usepackage{epsfig}
\usepackage{graphicx}
\usepackage{amsmath}
\usepackage{amssymb}
\usepackage{dsfont}
\usepackage{graphics}
\usepackage[utf8]{inputenc}
\usepackage{enumitem}
\usepackage[dvipsnames]{xcolor}
\newcommand\blfootnote[1]{%
  \begingroup
  \renewcommand\thefootnote{}\footnote{#1}%
  \addtocounter{footnote}{-1}%
  \endgroup
}
\usepackage[pagebackref=true,breaklinks=true,letterpaper=true,colorlinks,bookmarks=false]{hyperref}

\cvprfinalcopy % *** Uncomment this line for the final submission

\def\cvprPaperID{15} % *** Enter the CVPR Paper ID here

\ifcvprfinal\fi
\newcommand{\tar}{\emph{target} label}
\newcommand{\post}{\emph{post-attack} label}
\newcommand{\pre}{\emph{pre-attack} label}
\newcommand{\posts}{\emph{post-attack} labels}
\newcommand{\pres}{\emph{pre-attack} labels}
\newcommand\norm[1]{\left\lVert#1\right\rVert}

\author{Konda Reddy Mopuri*, Vaisakh Shaj*, R. Venkatesh Babu\\
Video Analytics Lab\\
Indian Institute of Science, Benguluru\\
{\tt\small kondamopuri@iisc.ac.in,vshaj@lincoln.ac.uk,venky@iisc.ac.in}
}
\title{Adversarial Fooling Beyond ``Flipping the Label"}

\begin{document}
\pagenumbering{gobble}
\maketitle

\begin{abstract}
Recent\blfootnote{* contributed equally} advancements in CNNs have shown remarkable achievements in various CV/AI applications. Though CNNs show near human or better than human performance in many critical tasks, they are quite vulnerable to adversarial attacks. These attacks are potentially dangerous in real-life deployments. Though there have been many adversarial attacks proposed in recent years, there is no proper way of quantifying the effectiveness of these attacks. As of today, mere fooling rate is used for measuring the susceptibility of the models, or the effectiveness of adversarial attacks. Fooling rate just considers label flipping and does not consider the cost of such flipping, for instance, in some deployments, flipping between two species of dogs may not be as severe as confusing a dog category with that of a vehicle. Therefore, the metric to quantify the vulnerability of the models should capture the severity of the flipping as well. In this work we first bring out the drawbacks of the existing evaluation and propose novel metrics to capture various aspects of the fooling. Further, for the first time, we present a comprehensive analysis of several important adversarial attacks over a set of distinct CNN architectures. We believe that the presented analysis brings valuable insights about the current adversarial attacks and the CNN models. 
\end{abstract}

%%%%%%%%% BODY TEXT
\section{Introduction}
\label{sec:intro}
Machine learning (ML) models are observed~(e.g.~\cite{evasion-mlkd-2013,prsystemsunderattack-pari-2014}) to be unstable to addition of structured noises known as \emph{adversarial perturbations}. These perturbations, despite being mild, tend to severely alter the inference of the ML models, generally referred to as \emph{fooling} the models. Over the time, number of different adversarial attacks (algorithms to fool) were proposed~(e.g.~\cite{intriguing-iclr-2014,deepfool-cvpr-2016,cw-sp-2017,fff-bmvc-2017,mopuri2018ask}) to demonstrate the vulnerability of the current ML systems, particularly the deep neural networks (DNNs). In case of a recognition model, it is understood that an adversarial attack is successful when the model predicts a different label upon adding the perturbation. Thus, all the existing works treat this \emph{label flipping} as \emph{fooling} the model. Therefore, they quantify the effectiveness of the underlying attack in terms of its \emph{fooling or success rate}, which is the percentage of successful flips.

However, fooling rate is a weak metric which fails to capture various important aspects of \emph{fooling} and ends up giving only partial picture about the attack or the target model. Specifically, it does not consider what the `post-attack' label is, and therefore fails to quantify the severity of the attack on either semantic or visual scale. Consequently, the fooling rate becomes apathetic to different flippings of the label and treats them identical. Though, from the robustness perspective, all label flippings should be treated equally, and a robust ML system should avoid any such susceptibility, in practice, different flippings (misclassifications) may inflict in different costs. For instance, in certain deployments, confusion between a pair of dog breeds is acceptable and not as severe as wrongly recognizing the stop sign on a highway.

% ######           Fig-1 (Different post-attack labels for googlenet) begin      #####################
\begin{figure*}[h]
\centering
\noindent\begin{minipage}{\textwidth}
\begin{minipage}{\textwidth}
  \centering
  \begin{minipage}{.12\textwidth}
  	\centering
    Input image\\\includegraphics[width=\linewidth]{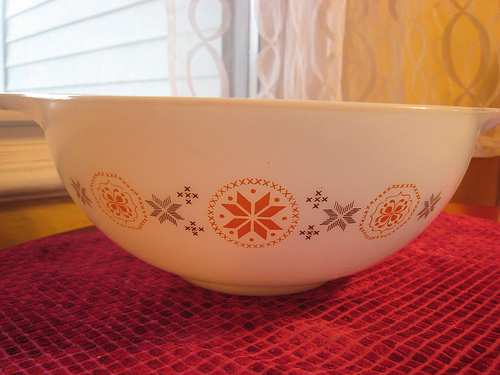}\\{\color{LimeGreen} Mixing-bowl}
  \end{minipage}
   \begin{minipage}{.12\textwidth}
   	\centering
    FGSM~\cite{explainingharnessing-iclr-2015}\\\includegraphics[width=\linewidth]{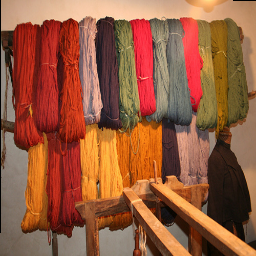}\\{\color{red}wool}
  \end{minipage}
  \begin{minipage}{.12\textwidth}
  	\centering
    iFGSM~\cite{physicalworld-arxiv-2016}\\\includegraphics[width=\linewidth]{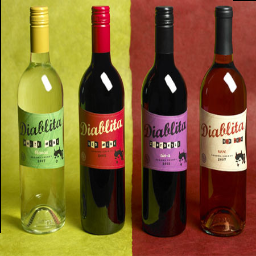}\\{\color{red}wine bottle}
  \end{minipage}
   \begin{minipage}{.12\textwidth}
   	\centering
    PGD~\cite{madry-iclr-2018}\\\includegraphics[width=\linewidth]{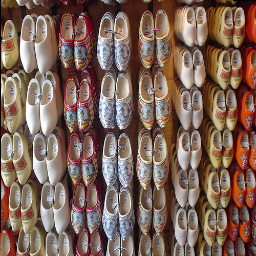}\\{\color{red}clog}
  \end{minipage}
   \begin{minipage}{.12\textwidth}
   	\centering
    DeepFool~\cite{deepfool-cvpr-2016}\\\includegraphics[width=\linewidth]{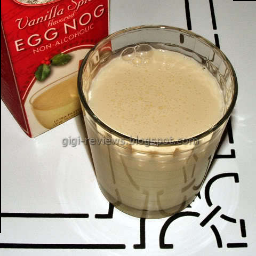}\\{\color{red}eggnog}
  \end{minipage}
  \begin{minipage}{.12\textwidth}
   	\centering
    UAP~\cite{universal-cvpr-2017}\\\includegraphics[width=\linewidth]{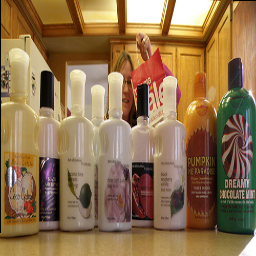}\\{\color{red}lotion}
  \end{minipage}
  \begin{minipage}{.12\textwidth}
   	\centering
    NAG~\cite{mopuri-cvpr-2018}\\\includegraphics[width=\linewidth]{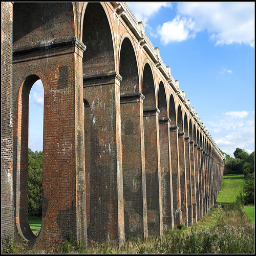}\\{\color{red}viaduct}
  \end{minipage}
  \begin{minipage}{.12\textwidth}
   	\centering
    GD-UAP~\cite{mopuri-pami-2018}\\\includegraphics[width=\linewidth]{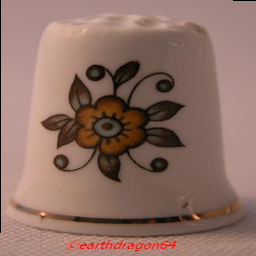}\\{\color{red}thimble}
  \end{minipage}
  \vspace{0.002\textwidth}
\end{minipage}
\vspace{0.002\textwidth}
\end{minipage}
\vspace{0.1cm}  
\caption{Fooling rate does not measure the extent of confusion caused by an adversarial attack. First column is the input image presented to GoogLeNet~\cite{googlenet-cvpr-2015} below which the \pre{} is mentioned in green. The subsequent columns show the representative images for the predicted \posts{} for $7$ different adversarial attacks. Note that the attacks are mentioned on top of the corresponding representative images taken from the ILSVRC dataset.
}
\label{fig:fooling-rate-shortcomings}
%\vspace{-0.5cm}
\end{figure*}
% ######           Fig-1 (Different post-attack labels for googlenet) end       ####################

Particularly, datasets with unwanted bias towards a set of semantically similar categories (e.g. ImageNet~\cite{imagenet-ijcv-2015} has $12\%$ dog categories) need sophisticated metrics for better analysis of the attacks and models. In such cases, weak metrics such as fooling rate could be misleading by providing only an incomplete picture of the models' vulnerability. Existing evaluation (e.g.~\cite{nips-advvision-2018,madry-iclr-2018,universal-cvpr-2017}) to compare various adversarial attacks is based solely on their fooling rate performance. However, the spectrum of existing attacks should be understood and analysed from not only the fooling rate perspective but also various other aspects of fooling, for instance, the actual semantic damage incurred due to the adversarial attack, etc. Moreover, fooling rate alone fails to bring out useful insights about the learning and the classification hyper-planes learned by these models. Therefore, the metric to quantify the effectiveness of the adversarial attacks should apprehend the severity of the flipping and provide better information about both the model and the underlying attack.

Hence, in this work, we present various important aspects of fooling caused by adversarial attacks. Specifically, we consider the Convolutional Neural Networks (CNN) trained for visual object recognition. %The popularity of these models due to the state of the art results in computer vision and our need for robust Artificial Intelligent (AI) models have motivated us towards the analysis presented in this work.
The major contributions of this work can be listed as:
\begin{itemize}[noitemsep,topsep=1pt]
    \item We present the shortcomings of the existing evaluation in order to emphasize the need for more sophisticated tools for analysis
    \item We propose a set of useful metrics (as baselines) to understand the attacks and models more comprehensively
    \item We present a detailed analysis and comparison of several important adversarial attacks over a set of distinct CNN architectures
\end{itemize}
The paper is organized as follows: section~\ref{sec:proposed} discusses the shortcomings in the existing evaluation and proposes multiple novel metrics to bring out the various important aspects of the adversarial fooling, section~\ref{sec:expts} presents comprehensive empirical analysis on several important image agnostic and image specific adversarial attacks, section~\ref{sec:disc} narrates some of the important observations, and finally section~\ref{sec:conclu} concludes the paper.
%-------------------------------------------------------------------------
\section{Fooling beyond \emph{Flipping}}
\label{sec:proposed}
In this section we demonstrate how fooling rate fails to capture various important aspects of confusing a DNN and present a set of useful metrics to better understand it.
\subsection{Shortcomings of Fooling Rate (FR)}
\label{subsec:shortcomings-fr}
Fooling rate, by definition is the percentage of success for an adversarial attack, i.e., expected number of times the attack is able to flip the label because of the added perturbation. For the rest of the paper, we define the label predicted on a clean image as \pre{} and that predicted on the corresponding adversarial image as \post{}. Thus, the fooling rate (FR) is defined as
\begin{equation}
    \frac{\sum_{i=1}^N \mathds{1}(pre{\text-}attack\\\:label(i) \neq post{\text-}attack \:label(i)) }{N}
\end{equation}
where $\mathds{1}$ is the indicator function that returns $1$ if the argument is true, else returns $0$, and $N$ is the total number of samples on which the attack is evaluated.

Clearly the fooling rate ignores what the \emph{post-attack} label is. All that it matters is, if it is different from the \emph{pre-attack} label or not. Thus, it does not measure the extent of the confusion caused by the adversarial attack. Since, some of the categories in the underlying dataset can be visually very close compared to others, not all mistakes are similar. For example, Figure~\ref{fig:fooling-rate-shortcomings} shows the \emph{post-attack} labels predicted by GoogLeNet~\cite{googlenet-cvpr-2015} for $7$ different adversarial attacks. The input image is shown in the first column, who's \emph{pre-attack} label is \emph{mixing-bowl}. Note that the subsequent columns show (hand-picked) representative images for the predicted \emph{post-attack} labels mentioned below them in red. Also, the corresponding adversarial attacks are mentioned above the representative images.

It is important to note that, though all the \posts{} are different from the \pre{}, visual patterns in some of the representative images are closer to the input image than others. For instance, \emph{eggnog} and \emph{thimble} will have bowl-like object patterns. On the other hand, \emph{wool} and \emph{viaduct} are visually very dissimilar to the \pre{}, mixing-bowl. However, fooling rate treats them all as successful fooling without considering the perceptual distance. In the following subsections, we present multiple metrics that reveal more information about the adversarial fooling.
\subsection{FR@K}
\label{subsec:fr@k}
We know that the fooling rate does not consider the rank of the \pre{} after the attack. However, it would be interesting to know how strong an attack can demote the \pre{} from rank-$1$, for instance, to compare different attacks or to understand the nature of the attack, etc. Therefore, we extend the definition of existing fooling rate in order to consider the rank of the \pre{} using ``Fooling rate at rank $K$" (FR@K). This means, for a given rank $K$, an attack is considered successful only if it assigns a rank $>K$ to the \pre{}. Therefore, after the attack, there will be at least $K\text{-}1$ other labels with greater confidence than the \pre. Intuitively, the $FR@K$ metric quantifies the extent of damage caused to the visual features discriminative to the \pre{} due to the attack. Thus, $FR@K$ is defined as

{\footnotesize
\begin{equation}
    \frac{\sum_{i=1}^N \mathds{1}(pre{\text-}attack\\\:label(i) \not\in \{top\text{-}K\: post{\text-}attack \:labels(i)\}) }{N}
\end{equation}
}
Note that when $K\text{=}1$, FR@K becomes the fooling rate. A similar metric in spirit has been proposed by Ganeshan \textit{et al.}~\cite{fda-iccv-2019} where they quantify the shifts in the ranks of \textit{pre} and \posts{} during the attack.
\subsection{Mean semantic confusion: QI-Wup}
\label{subsec:qiwup}
Existing evaluation completely ignores to measure the ``semantic damage'' caused by the adversarial attacks. This is because the fooling rate is apathetic to different flippings by adversarial attacks. However, there exist various attacks that are designed with very different objective functions though the ultimate goal is to fool the target model. Thus, it is quite possible that a given model incurs varying levels of confusion for different adversarial attacks. For instance, in Figure~\ref{fig:fooling-rate-shortcomings}, GoogLeNet confuses \emph{Mixing-bowl} to a range of different labels from \emph{wine bottle} to \emph{viaduct}. Note that the \posts{} resulted by different attacks lie at different semantic distance to the \pre.  Also, different deployment environments (e.g. household, commercial, military, etc.) would work with varied levels of acceptable confusion. It is beneficial to have a useful metric that can quantify the actual semantic damage incurred by a given model for various attacks.

In this subsection we introduce an intuitive metric named ``Mean semantic confusion'', that can quantify the semantic damage caused by an adversarial attack (or in other words, the semantic damage incurred by a given model). We adapt the familiar word similarity metrics such as Wu-Palmer~\cite{wu-palmer-acl-94} to measure the severity of the flipping on a semantic scale. We define the Quantized Inverse Wup similarity (QI-Wup) as 

{\scriptsize
\begin{equation}
    QI\text{-}Wup=\begin{cases}
    1, & \text{if $Wup(\pre,\post)<T_s$}.\\
    0, & \text{otherwise}.
  \end{cases}
\end{equation}
}
where $Wup(x,y)$ is the Wup similarity between the words $x$ and $y$ and $T_s$ is a threshold chosen based on the target deployment environment. %Since the Wup similarity lies in $[0, 1]$, the $QI\text{-}Wup$ also lies in $[0, 1]$. 
Note that the proposed $QI\text{-}Wup$ metric deems a \emph{flipping} as \emph{fooling} only when the semantic similarity is less than an acceptable threshold $T_s$. Different thresholds can be chosen for various deployment scenarios~(see supplementary materials) based on the acceptable semantic confusion.

Therefore, the mean semantic confusion can be computed as the average $QI\text{-}Wup$ score over a set of evaluation samples. Note that Wup measure is one choice of word (semantic) similarity and we can chose any other similarity.  

\subsection{Mean visual confusion: QI-Vis}
\label{subsec:qivis}
Another very important aspect of adversarial fooling is to understand if the CNN models get confused only among ``visually'' closer labels or even the dissimilar ones (refer to section~\ref{subsec:vis-sem}). For instance, as shown in Figure~\ref{fig:fooling-rate-shortcomings}, GoogLeNet gets confused to recognize the \emph{Mixing-bowl} to various labels. The Deepfool~\cite{deepfool-cvpr-2016} attack flips the label to \emph{eggnog}. Note that 
the eggnog images always have a bowl like container to hold it. Therefore, in this case the attack tries to fool the model to predict another class that has similar visual patterns. While in case of other attacks, particularly the image-agnostic attacks such as UAP~\cite{universal-cvpr-2017}, GD-UAP~\cite{mopuri-pami-2018} it is less observed. 

In case of measuring the semantic damage, we have a hierarchical structure (graph) such as WordNet~\cite{wordnet-95} to understand how the labels are semantically related. However, such a data structure for `visual' relations does not exist. It is very difficult to collect the visual similarities among the categories via human annotations given large number of classes and intra class variations. Thus, (similar to~\cite{zskd-icml-2019}) we collect these visual similarities from the learned model itself. The final layer of any classification layer will be a fully connected $(fc)$ layer with a softmax nonlinearity. Each neuron in this layer corresponds to a class $(c)$ and its activation is treated as the confidence/probability $(S_c)$ predicted by the model to that class. The weights connecting previous layer to this neuron $(W_c)$ can be considered as the template of the class $(c)$ learned by the network. This is because, the confidence predicted $(S_c)$ is proportional to the alignment of the previous layer's output with the template $(W_c)$. It becomes maximum when the previous layer's output is a positive scaled version of this template $(W_c)$. On the other hand, if the output of the previous layer is misaligned with the template $W_c$, the confidence $S_c$ is reduced.

Therefore, we compute the visual similarity (as perceived by the target model) between a pair of classes $i$ and $j$ as

\begin{equation}
    Vis(i,j)=\frac{{W_i}^T{W_j}}{ \norm{W_i} \norm{W_j} }
\end{equation}

Using this similarity score, we define the Quantized Inverse Visual Similarity $(QI\text{-}Vis)$ as

{\scriptsize
\begin{equation}
    QI\text{-}Vis=\begin{cases}
    1, & \text{if $Vis(\text{\pre{}},\text{\post{}})<T_v$}.\\
    0, & \text{otherwise}.
  \end{cases}
\end{equation}
}
Where $T_v$ is a threshold chosen in order to impose a desired tolerance in terms of visual confusion. Note that $Vis(i,j)$ may not lie in $[0, 1]$ unlike the $Wup$ measure. Also, $Vis$ is one of the possible visual similarity measures, and $QI\text{-}Vis$ can be computed over any such measure. Further, for a given dataset, our visual similarities are model specific, however it is still a valid candidate for visual similarity since we observe that multiple models closely agree upon these similarities. For the $4$ models considered in our experiments, the variance of class similarities computed across the $1000$ ILSVRC categories is very small with a mean value of $6 \times 10^{-4} (\pm 5.95 \times 10^{-4})$. See supplementary materials for more details.
%------------------------------------------------------------------------
% ######           Fig-2 (FR@K experimental results) begin       ####################
\begin{figure*}[ht]
%\centering
%\noindent\begin{minipage}{\textwidth}
  \begin{minipage}{.24\textwidth}
  \centering
    \includegraphics[width=\linewidth]{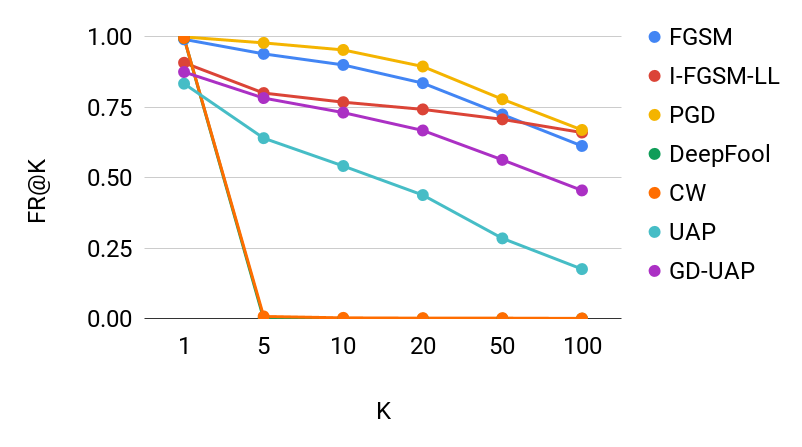}\\CaffeNet
  \end{minipage}
   \begin{minipage}{.24\textwidth}
   \centering
    \includegraphics[width=\linewidth]{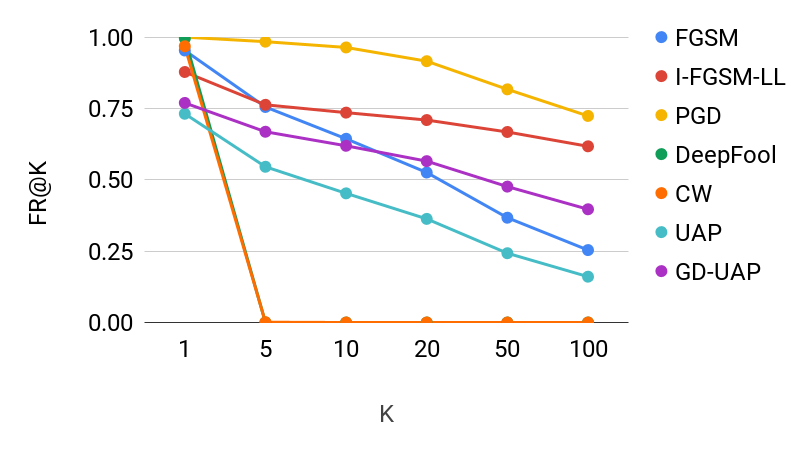}\\GoogLeNet
  \end{minipage}
  \begin{minipage}{.24\textwidth}
  	\centering
    \includegraphics[width=\linewidth]{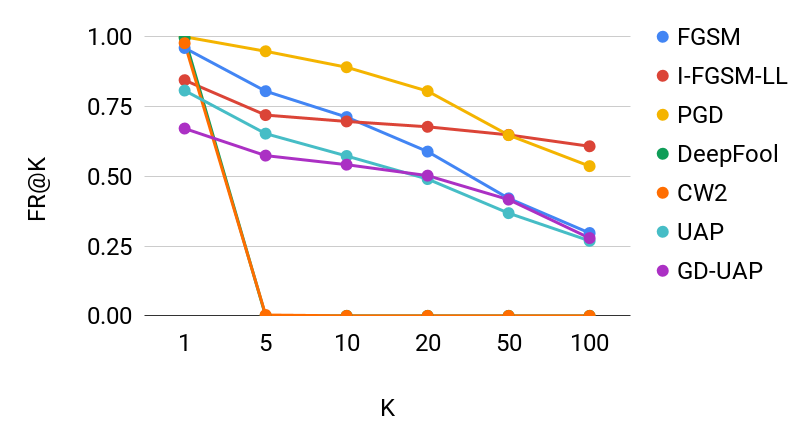}\\VGG$-19$
  \end{minipage}
   \begin{minipage}{.24\textwidth}
   	\centering
    \includegraphics[width=\linewidth]{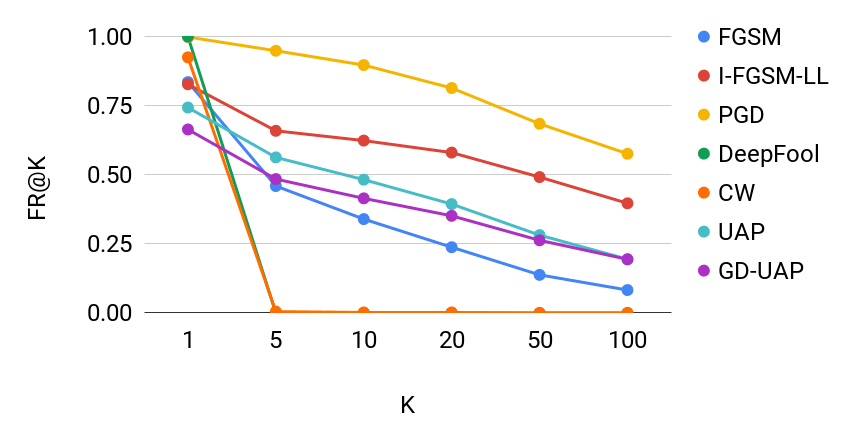}\\ResNet$-152$
  \end{minipage}

\iffalse  
\noindent\begin{minipage}{\textwidth}
  \centering
  \begin{minipage}{.48\textwidth}
  	\centering
    \includegraphics[width=\linewidth]{figures_aaai/fr@k/vgg-19.png}\\VGG$-19$
  \end{minipage}
   \begin{minipage}{.48\textwidth}
   	\centering
    \includegraphics[width=\linewidth]{figures_aaai/fr@k/resnet-152.png}\\ResNet$-152$
  \end{minipage}
  \vspace{0.002\textwidth}
\end{minipage}
\fi
\caption{FR@K computed for various CNN models for multiple adversarial attacks. Note that the attacks are mentioned in the legend and the model name is provided below the corresponding plot.}
\label{fig:fr@k-expts}
%\vspace{-0.5cm}
\end{figure*}
% ######           Fig-2 (FR@K experimental results) end       ####################
\section{Experiments}
\label{sec:expts}
In this section we present the experimental analysis to show the effectiveness of the aforementioned metrics.
We performed all our experiments on the models trained for object recognition on ILSVRC~\cite{imagenet-ijcv-2015} dataset. To be comprehensive, we considered models from different architecture families, namely, CaffeNet~\cite{deepcnn-nips-2012}, GoogLeNet~\cite{googlenet-cvpr-2015}, VGG-$19$~\cite{vgg-iclr-2015}, ResNet-$152$~\cite{he2016deep}. Note that the evaluation is performed on $10000$ correctly classified images from the validation set. We considered a range of adversarial attacks that include image specific, image agnostic and iterative variants. Specifically, we evaluated on the following
\begin{itemize}[noitemsep]
    \item Fast Gradient Sign Method (FGSM)~\cite{explainingharnessing-iclr-2015}
    %\item Iterative FGSM (I-FGSM)
    \item Projected Gradient Descent (PGD)~\cite{madry-iclr-2018}
    \item DeepFool~\cite{deepfool-cvpr-2016}
    %\item Iterative FGSM least likely (I-FGSM-LL)~\cite{explainingharnessing-iclr-2015}
    \item Carlini and Wagner (CW)~\cite{cw-sp-2017}
    \item Univeral Adversarial Perturbations (UAP)~\cite{universal-cvpr-2017}
%item Network for Adversary Generation (NAG)~\cite{mopuri-cvpr-2018}
    \item Generalizable Data-free UAP (GD-UAP)~\cite{mopuri-pami-2018}
\end{itemize}
We briefly introduce these attacks along with the required notation for the ease of reference.
\begin{itemize}[noitemsep]
\item $X:$ clean image from the dataset.
\item $X^{adv}:$ potential adversarial image crafted from $X$.
\item $y_{true}:$ ground truth label corresponding to $X$.
\item $y_{pred}:$ prediction of the neural network for $X$.
\item $f:$ mapping function that represents neural network
\item $\epsilon:$ strength of perturbation added to the clean image.
%\item $\theta:$ parameters of the neural network.
\item $J:$ loss function used to train the neural network.
\item $\nabla J:$ gradient of the loss $J$ with respect to image $X$. %_{x}
\end{itemize}
\textbf{Fast Gradient Sign Method} (FGSM): is a simple way to craft adversaries. They linearly approximate the loss function and compute the gradient as the adversarial direction to perturb the input:
\begin{equation}
    X^{adv}=X+\epsilon . sign(\nabla J(X,y_{true}))
\end{equation}
\\
\textbf{I-FGSM-LL} is a variety of the FGSM attack, in which we iteratively (with small steps) compute the perturbation in order to decrease the loss for predicting a `least-likely' label.\\
\\
\textbf{Projected Gradient Descent} (PGD): One can think of FGSM attack as a single-step scheme to maximize the loss function within the $\epsilon$ ball around $X$, which is represented by $\mathcal{S}$. A powerful attack would be an iterative variation, $FGSM^{k}$, which is essentially performing Projected Gradient Descent on the negative loss function:
\begin{equation}
    X^{t+1} = \Pi_{X+\mathcal{S}} (X^t + \alpha . sign(\nabla J(X,y_{true})))
\end{equation}
where $t$ is the iteration, and $\alpha$ is the maximum perturbation at each iteration.\\
\\
\textbf{DeepFool}: defines an adversarial perturbation as the minimal perturbation $v$ that is sufficient to change the inference of the classifier:
\begin{equation}
    min_{v} \text{\:\:subject to\:\:} f(X+v) \neq f(X)
\end{equation}
Their algorithm is a greedy method that approximates the non-linear class boundaries as hyper-planes and in practice (generally) yields a small and effective perturbation.\\
\\
\textbf{Carlini and Wagner attack} (CW): makes the
perturbations quasi-imperceptible by minimizing the $l_p$ norm. This is achieved by solving the following optimization problem:
\begin{align}
    \text{minimize\:\:} & \lvert\lvert v \rvert \rvert_p + c. g(X+v) \\
    \text{such that\:\:} & X+v \in [0,1]
    \label{eqn:cw}
\end{align}
Here $g$ is a surrogate objective function such that $g(x + v) \leq 0$ $\iff$ $f(x + v) = t$. Here $v$ is the adversarial perturbation, $c$ is a constant, $t$ is the target label and $f$ is a mapping function that represents the neural network.\\
\\
\textbf{Universal Adversarial Perturbation} (UAP): presented an algorithm to compute a perturbation agnostic to the input samples, known as `Universal' Adversarial Perturbation $(v)$. 
\begin{align}
    \lvert \lvert v \rvert \rvert_p & \leq \epsilon \\ 
    \mathcal{P}_{X \sim \mu} (f(X+v)\neq & f(X)) \geq 1-Th
    \label{eqn:uap}
\end{align}
where, $Th$ quantifies the desired fooling rate. Their algorithm simply accumulates the individual sample specific DeepFool~\cite{deepfool-cvpr-2016} adversarial perturbations and regularly projects into the feasible ball of the perturbations.\\
\\
\textbf{Generalized Data-free UAP} (GD-UAP): presented a data-free `activation' loss that is generalizable across various vision tasks to compute a UAP. They attempt achieve an objective similar to eqn.~(\ref{eqn:uap}) without utilizing any data samples via optimizing the following loss:
\begin{align}
    Loss = & -\log(\prod_{i=1}^{K} \lvert \lvert l_i(v) \rvert \rvert_2) \\ \text{such that\:\:} & \lvert \lvert v \rvert \rvert_p \leq \epsilon
\end{align}
where, $l_i(v)$ is the response of $i^{th}$ layer, and $K$ is the total number of layers in the model.

We chose the best hyper-parameters for all the attacks (e.g. number of iterations, etc.) as mentioned in the corresponding works or via conducting ablation. Also, we consider $l_\infty$ norm of $10$ for restricting the strength of the perturbation $(\epsilon)$. However, note that the CW and DeepFool attacks because of their nature, do not impose the same max-norm restriction on the perturbations. In case of CW attack, we use a relatively large value of 15 for binary-search steps, which helps in determining the trade off-constant $c$.

In summary, we consider the best operating parameters for all the attacks in order to ensure the comparison is fair. We present some useful ablations over these hyper-parameters (e.g. CW) in the supplementary material.
% ######           Fig-3 (QI-Wup experimental results) begin       ####################
\begin{figure*}[t]
%\centering
%\noindent\begin{minipage}{\textwidth}
  \begin{minipage}{.24\textwidth}
  \centering
    \includegraphics[width=\textwidth]{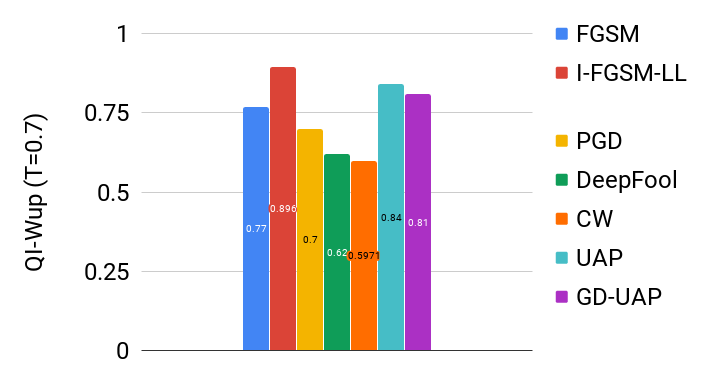}\\CaffeNet
  \end{minipage}
   \begin{minipage}{.24\textwidth}
   \centering
    \includegraphics[width=\textwidth]{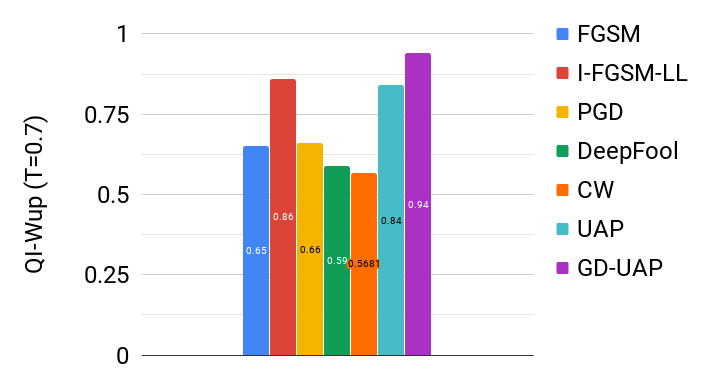}\\GoogLeNet
  \end{minipage}
  \begin{minipage}{.24\textwidth}
  	\centering
    \includegraphics[width=\textwidth]{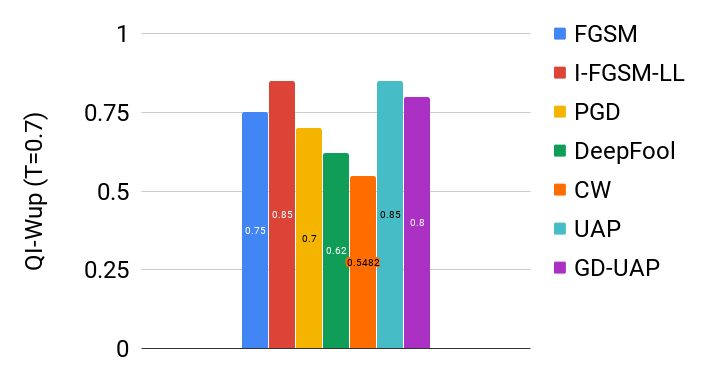}\\VGG$-19$
  \end{minipage}
   \begin{minipage}{.24\textwidth}
   	\centering
    \includegraphics[width=\textwidth]{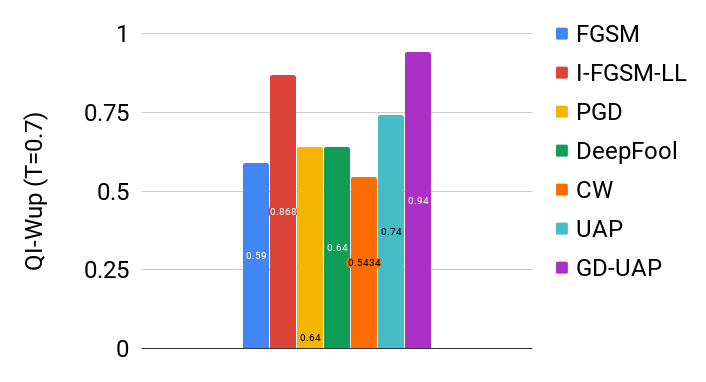}\\ResNet$-152$
  \end{minipage}

\caption{QI-Wup computed for various CNN models for multiple adversarial attacks. Note that the attacks are mentioned in the legend and the model names are provided below the corresponding plot.}
\label{fig:qiwup-expts}
%\vspace{-0.5cm}
\end{figure*}
% ######           Fig-3 (QI-Wup experimental results) end       ####################
\subsection{FR@K}
\label{subsec:fr@k-expts}
Fooling rate at rank K $(FR@K)$ gives the success rate of an attack to demote the \pre{} beyond the first $K$ ranks after the attack. 
Figure~\ref{fig:fr@k-expts} shows $FR@K$ for various attacks computed on multiple models. We computed the results for $K={1, 2, 5, 10, 20, 50, \text{ and}, 100}$. Note that $K$ can take a maximum value of $999$ since the total number of categories in ILSVRC is $1000$. We can notice that, as expected, the $FR@K$ falls with $K$, since it gets more difficult for the attack to demote the \pre{} further.% Also, the decay trend is observed to be consistent to a great extent with all models.

However, we make a very important observation about the DeepFool~\cite{deepfool-cvpr-2016} and CW~\cite{cw-sp-2017} attacks. They could not demote the \pre{} further. $FR@K$ becomes zero for all the higher values of $K$. Specifically, in all the $10000$ correctly classified validation images that we considered for evaluation, the highest rank DeepFool could successfully demote to is $3$. In almost all the cases, it makes the model to simply swap the first and second labels. Note that this behaviour is consistent across all the models. This behaviour can be explained from the design of the attack. The DeepFool algorithm searches for the nearest decision boundary to the input sample and finds a contamination in order to move the sample across that boundary. Therefore, it ends up fooling the model to predict the nearest class to the \pre{} which is top-$2$ (or top-$3$) label before the attack. For the CW attack, we used a variant in which the adversary is crafted to make the \pre{} least-likely (i.e. poorly ranked after the attack). We observed that the resulting attack's behaviour is similar to the `best case' scenario proposed by the authors in which the \post{} is the second most probable label before perturbing. Please note that the CW attack doesn't guarantee the $l_p$ norm of the perturbation to be smaller than a predefined value, rather, it only minimises through the objective (eq. 9). Hence, we chose a variant whose average perturbation is comparable to the other attacks in order to provide a fair comparison. Please refer to the supplemental material for a stronger version of the CW attack.

Apart from extracting such hidden details about the attacks, we can also convert these graphs into metrics such as the area under the curve (AuC), which can be used for a direct quantitative comparison. For instance, if one looks for an attack that can demote the \pre{} strongly, it should have a high AuC for the $FR@K$ plot. Similarly a robust model would have a low AuC for multiple attacks. Table~\ref{tab:auc} shows the AuC computed for the curves shown in Figure~\ref{fig:fr@k-expts}. In terms of performance with respect to AuC metric, attacks such as PGD and I-FGSM-LL inflict the maximum disruption to the visual features discriminative to the \pre{}. Notice that the AuC generally decreases as the models get sophisticated from left to right. This increased robustness can be attributed to the advanced network architectures with efficient regularizers such as dropout and batchnorm.% Note that even the ResNet-$152$ architecture incurs a very high AuC for iterative attacks such as CW and PGD.
\begin{table}[]
\centering
{\small
\begin{tabular}{|l|c|c|c|c|} \hline
          & \scriptsize{CaffeNet} & \scriptsize{GoogLeNet} & \scriptsize{VGG-19} & \scriptsize{ResNet-152} \\ \hline
FGSM      & 0.7      & 0.4       & 0.44   & 0.19       \\ \hline
%I-FGSM    & \textbf{0.78}     & \textbf{0.82}      & \textbf{0.69}   & 0.73       \\ \hline
\small{IFGSM-LL} & 0.66     & 0.61      & 0.59   & 0.43       \\ \hline
PGD       & \textbf{0.79}     & \textbf{0.83}      & \textbf{0.68}   & \textbf{0.71}        \\ \hline
DeepFool  & 0.02     & 0.02      & 0.02   & 0.02       \\ \hline
CW        & 0.02      & 0.02      & 0.02   & 0.02       \\ \hline
UAP       & 0.33     & 0.28      & 0.4    & 0.32       \\ \hline
%NAG       & 0.53     & 0.5       & 0.4    & 0.3        \\ \hline
GD-UAP    & 0.58     & 0.5       & 0.42   & 0.29      \\ \hline
\end{tabular}
}
\caption{Area under the $FR@K$ curves for multiple models under various adversarial attacks.}
\label{tab:auc}
%\vspace{-0.5cm}
\end{table}
\subsection{QI-Wup}
\label{subsec:qiwup-expts}
%AI systems are being deployed in the real world and soon more such systems are going to interact with humans and among themselves. Given their susceptibility to adversarial attacks, it is required to have a better understanding of the same in terms of the `semantics'. Particularly in case of the vision models, it is essential to have metrics to measure the damage caused by the adversarial attacks. Fooling rate can not compare two attacks on a semantic 
Figure~\ref{fig:qiwup-expts} presents the $QI\text{-}Wup$ measure computed for multiple models over various attacks. For all these experiments we have used a threshold $(T_s)$ of $0.7$ on the Wup similarity.% Note that the attacks are mentioned in the legend and the model names are provided below the plots.

It can be observed that image agnostic attacks such as UAP~\cite{universal-cvpr-2017}, GD-UAP~\cite{mopuri-pami-2018} consistently result in a strong semantic damage compared to the image specific counterparts. This can also be explained from the fooling patterns of these attacks. In case of image agnostic attacks, the existence of `dominant' \posts{} is observed. That is, after the attack, the \pres{} are generally mapped to a small set of sink classes. For instance, in case of UAP, \posts{} computed for all the $50000$ validation images on GoogLeNet comprise only $17\%$ of the total categories. It is hypothesized~\cite{universal-cvpr-2017} that the dominant labels occupy large space and hence represent good candidates for these attacks to fool the models. However, this is not the case with the image specific attacks. It is observed that the DeepFool and CW attacks generally inflict the smallest semantic damage compared to the others. Note that this is consistent with the observation in section~\ref{subsec:fr@k-expts} that they result in least $FR@K$.

Among the image specific attacks, I-FGSM-LL inflicts maximum semantic confusion. This is understandable, since this attempts to make the model predict the least-likely label which is generally (visually) far away from the \pre{}. Hence, in general, the two labels should also be semantically very far away. However, it is interesting to note that PGD and FGSM attacks also inflict stronger semantic damage.

\begin{figure}
    \centering
    \includegraphics[width=0.48\textwidth]{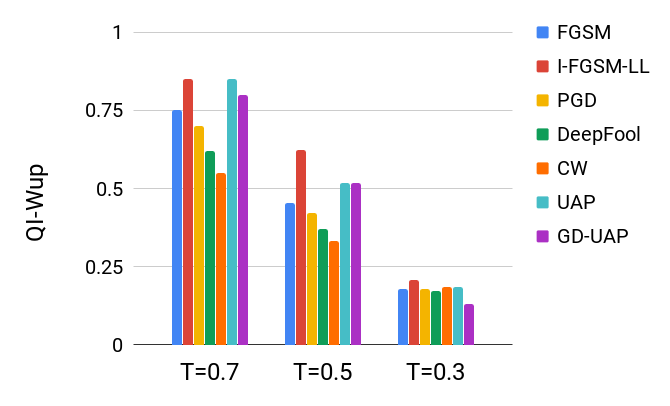} %-2 for iccv
    \caption{Mean semantic confusion (QI-Wup) caused by various attacks on VGG-$19$ computed with different threshold values for Wup similarity.}
    \label{fig:qiwup-ablation}
    %\vspace{-0.5cm}
\end{figure}

We experimented with different threshold values $(T_s)$ for the Wup measure. Figure~\ref{fig:qiwup-ablation} shows the $QI\text{-}Wup$ values for different threshold values computed for the VGG-$19$ model. Note that as threshold value decreases it becomes difficult for the attacks to cause severe semantic confusion and the metric becomes close to indiscernible. On the other hand, higher value of $T_s$ will bring out the subtle differences among the attacks. %Also, note that Wup measure is one choice and the metric can be computed with any suitable semantic similarity measure.
% ######           Fig-4 (QI-Wup ablation results) begin       ####################
% ######           Fig-4 (QI-Wup ablation results) end       ####################
% ######           Fig-5 (QI-Vis experimental results) begin       ####################
\begin{figure*}[h]
%\centering
%\noindent\begin{minipage}{\textwidth}
  \begin{minipage}{.24\textwidth}
  \centering
    \includegraphics[width=\linewidth]{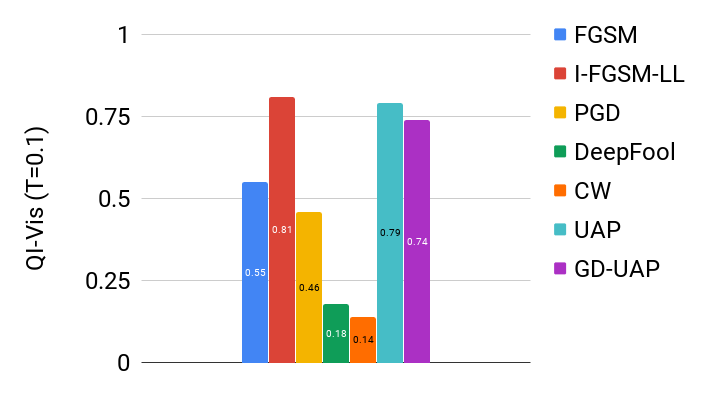}\\CaffeNet
  \end{minipage}
   \begin{minipage}{.24\textwidth}
   \centering
    \includegraphics[width=\linewidth]{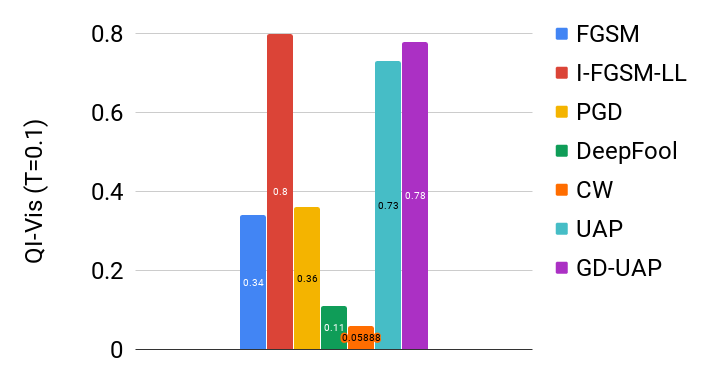}\\GoogLeNet
  \end{minipage}
    %\vspace{0.002\textwidth}
    %\end{minipage}
\begin{minipage}{.24\textwidth}
  	\centering
    \includegraphics[width=\linewidth]{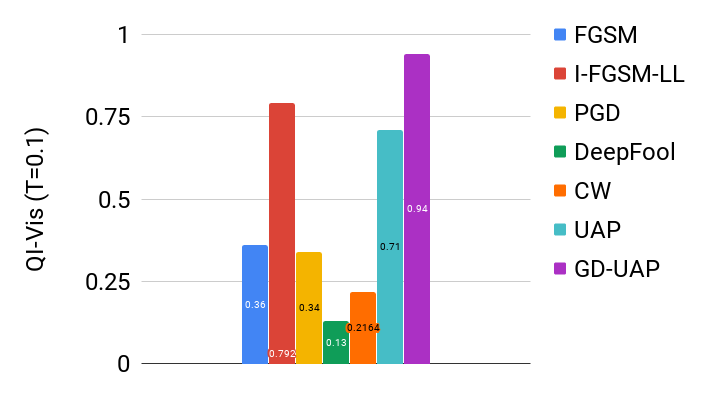}\\VGG$-19$
  \end{minipage}
   \begin{minipage}{.24\textwidth}
   	\centering
    \includegraphics[width=\linewidth]{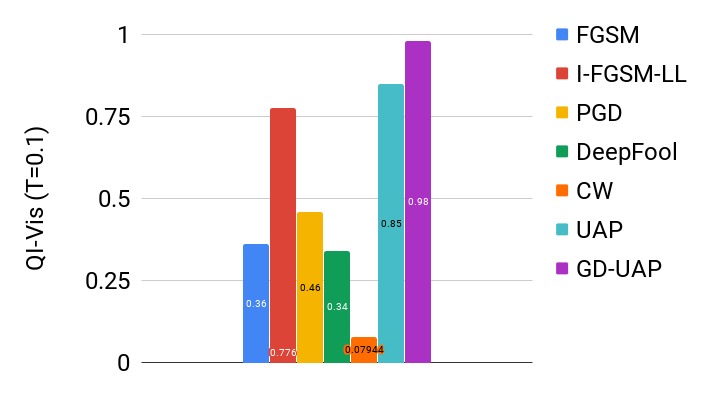}\\ResNet$-152$
  \end{minipage}
  
\iffalse

\noindent\begin{minipage}{\textwidth}
  \centering
  \begin{minipage}{.48\textwidth}
  	\centering
    \includegraphics[width=\linewidth]{figures_aaai/qivis/vgg-19.png}\\VGG$-19$
  \end{minipage}
   \begin{minipage}{.48\textwidth}
   	\centering
    \includegraphics[width=\linewidth]{figures_aaai/qivis/resnet-152.png}\\ResNet$-152$
  \end{minipage}
  %\vspace{0.002\textwidth}
  \end{minipage}
 \fi 
  
\caption{QI-Vis computed for various CNN models for multiple adversarial attacks. Note that the model is mentioned below the corresponding plot.}
\label{fig:qivis-expts}
\end{figure*}
% ######           Fig-5 (QI-Vis experimental results) end       ####################
% ######           Fig-6 (Interesting examples) begin       ####################
\begin{figure}[t]
    \centering
    \includegraphics[width=0.45\textwidth]{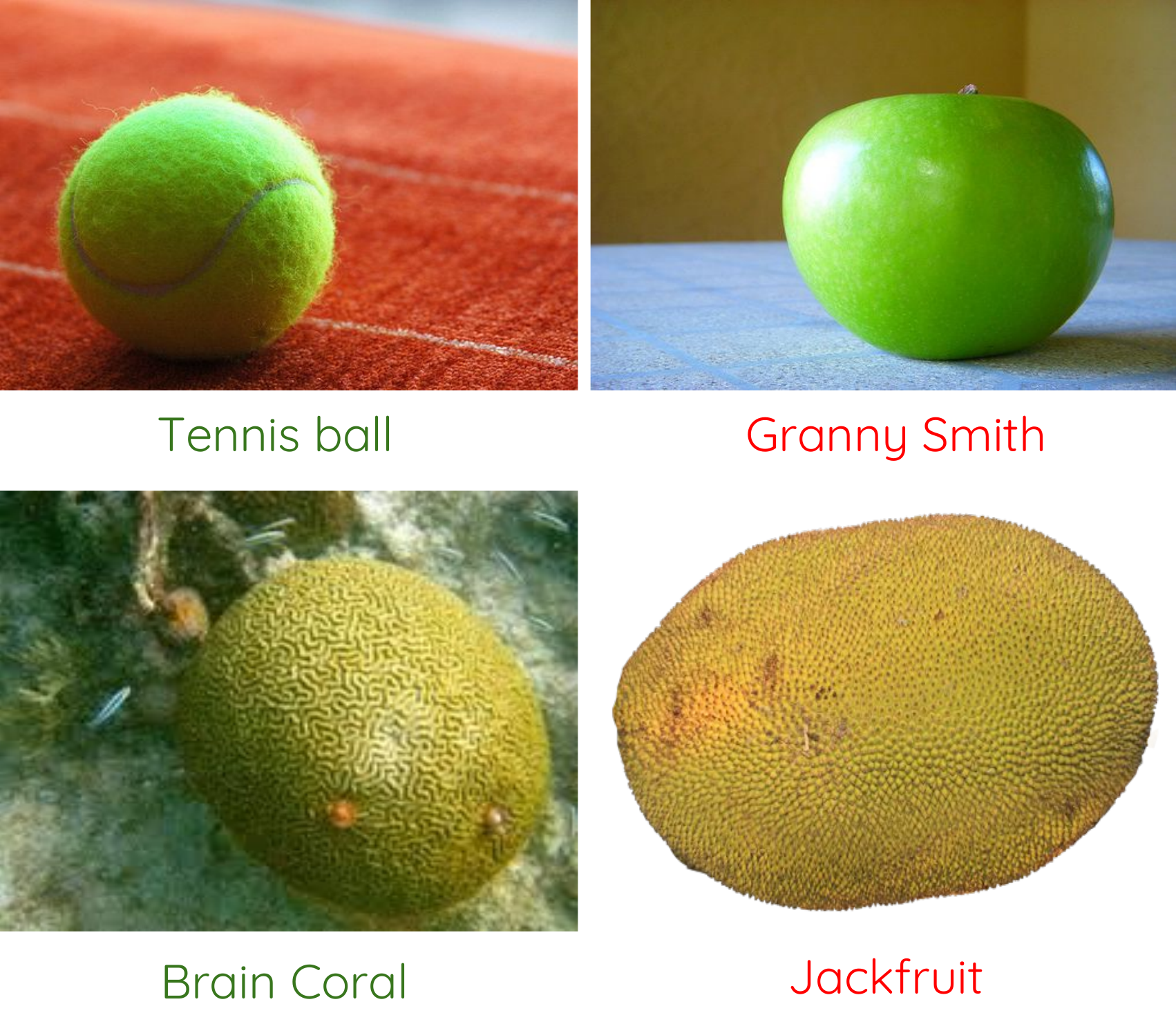}
    \caption{Similar ‘object’ patterns can cause severe semantic confusion.
First column shows input samples and their \pres{} and the second column shows the representative images from
the predicted \posts{}.}
    \label{fig:vis-sem}
    %\vspace{-0.5cm}
\end{figure}
% ######           Fig-6 (Interesting examples) end       ####################
\subsection{QI-Vis}
\label{subsec:qivis-expts}
%Understanding the process of adversarial fooling in terms of the visual confusion is another important aspect.
%In this section, we present our observations in terms of the visual dissimilarity between the \emph{pre} and \posts{} as discussed in section~\ref{subsec:qivis}.
Figure~\ref{fig:qivis-expts} shows the $QI\text{-}Vis$ metric computed for various adversarial attacks over multiple CNN models. Note that we have used visual similarities extracted from the corresponding CNN models as detailed in section~\ref{subsec:qivis} and a threshold $(T_v)$ value of $0.1$. That means, a flipping is considered `visually fooling' only if the visual similarity between the \pre{} and \post{} is less than $0.1$. Note that this threshold is very small compared to $T_s$. This is because the Wup similarity scores in general are high even for a trivial case of semantically dissimilar labels. For e.g. \emph{Brain Coral} and \emph{Jack fruit} (Figure~\ref{fig:vis-sem}) have a Wup similarity of $0.46$. On the other hand, the visual similarities computed from the network (sec.~\ref{subsec:qivis}) are very low. Supplemental document provides the percentile graph computed for GoogLeNet visual similarities. We observe that $95\%$ of the similarities are less than $0.1$. Thus, we chose $T_v=0.1$ for our analysis.

Due to the presence of dominant labels in the \posts{}, the image agnostic attacks (UAP, GD-UAP) cause more visual confusion compared to the image specific attacks. Image specific attacks such I-FGSM-LL also inflict significant visual confusion to the models. Also, note that the visual confusion caused by FGSM and PGD is significantly higher than that by CW and DeepFool.%ery high values of $QI\text{-}Vis$ are a result of the threshold $T=0.1$. 

In summary, iterative attacks remove most evidence for the \pre{} and cause severe demotion. Image agnostic, and targeted attacks can do severe damage with respect to all the three metrics on most of the CNN models. Interestingly, simple FGSM based attacks also cause significant damage (which is non-trivial given only their fooling rate performance) with respect to proposed metrics. 

Among the models, ResNet-$152$ has the least AuC followed by VGG-$19$. Further, ResNet-$152$ has smallest $QI\text{-}Wup$ followed by GoogLeNet. In terms of $QI\text{-}vis$, GoogLeNet incurs least damage followed by VGG-$19$. Interestingly, ResNet-$152$ demonstrates the highest $QI\text{-}Vis$.
%-------------------------------------------------------------------------
% ######           Fig-7 (Interesting examples) begin       ####################
\begin{figure}[t]
    \centering
    \includegraphics[width=0.45\textwidth]{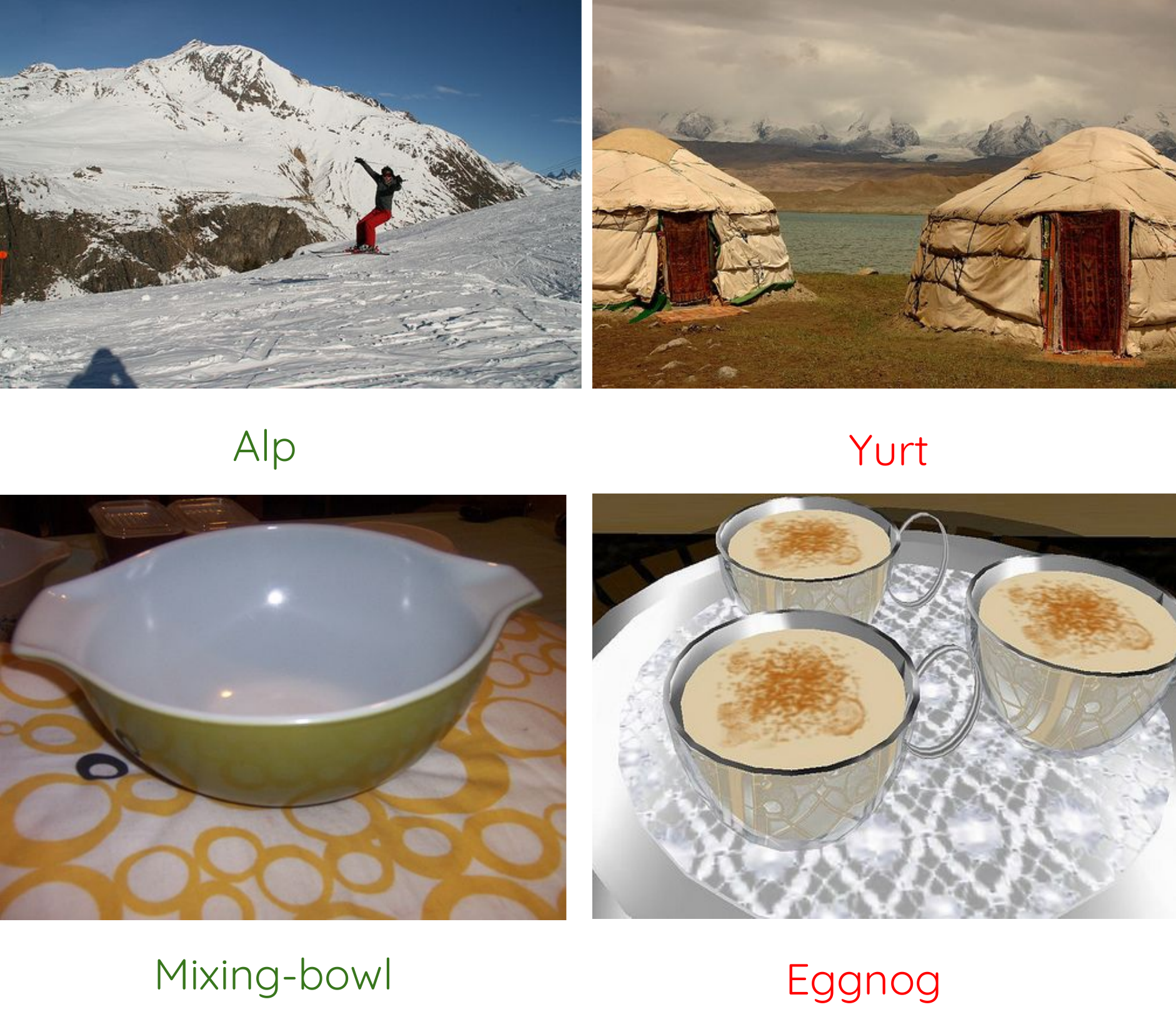}
    \caption{Similar context can also cause models to easily confuse across classes. First column shows input samples and the corresponding \pres{} and the second column shows the representative images from the predicted \posts{}.}
    \label{fig:context}
\end{figure}
% ######           Fig-7 (Interesting examples) end       ####################
%\vspace{-0.25cm}
\section{Discussion}
\label{sec:disc}
In this section, we present some interesting observations
that we have come across while analysing the attacks.
\subsection{Visual vs Semantic similarity}
\label{subsec:vis-sem}
It is as challenging as it is interesting to discuss `if the visual similarity correlates to the semantic similarity' or vice versa. In computer vision, it is often taken for granted that they both are correlated~\cite{vis-sem-cvpr-2011}. While analysing the confusion of the CNN models to adversarial attacks, we have come across interesting examples about the visual and semantic similarity. Figure~\ref{fig:vis-sem} shows a pair of example images presented to a trained GoogLeNet under the DeepFool adversarial attack. The first column shows the original images with the ground truth label mentioned in green below them. In the second column we show the representative samples for the \posts{} predicted by the model. Note that the \posts{} are mentioned below them in red.

We intentionally chose the DeepFool attack for it causes the least semantic confusion. However, in this case it is very clear that the \posts{} are semantically far away (Wup similarities are $0.35$ and $0.46$) from the \pres{}. However, upon investigating, we found that the visual patterns of the \posts{} are similar to those of the \pres{}. Their visual similarities given by the model (refer to sec.~\ref{subsec:qivis}) are $0.17$ and $0.23$ respectively. Please note that they are large compared to the $95\%$ percentile similarity of $0.1$. In both the cases, the \posts{} are ranked $2$ before the attack. It can be explained with the learning procedure. The only input that the CNN model has received about the object categories is the images belonging to those categories. Therefore, the model tries to learn the discriminative visual patterns for each category from the corresponding samples. Visually similar patterns thus can cause the model to confuse across categories that are semantically far apart, though such cases are not very common. In case of the examples presented in Figure~\ref{fig:vis-sem}, due to the similar visual patterns DeepFool attack could successfully inflict a severe semantic confusion. However, in general we observe the top ranked labels to be both visually and semantically similar.
% ######           Fig-8 (Dog breeds) begin       ####################
\begin{figure}[t]
    \centering
    \includegraphics[width=0.45\textwidth]{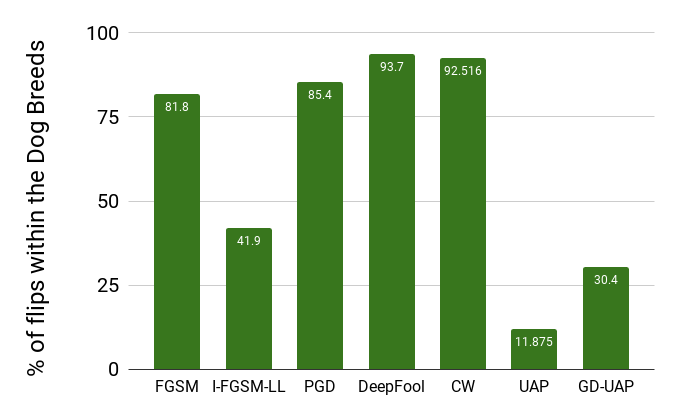}
    \caption{Confusion among fine-grained categories. Figure shows the $\%$ of dog samples that are misclassified as another dog category. Note that the analysis is performed on GoogLeNet for $800$ validation images from $117$ dog categories of ILSVRC dataset.
    }
    \label{fig:dog-breeds}
    %\vspace{-0.5cm}
\end{figure}
% ######           Fig-8 (Dog breeds) end       ####################
\subsection{Influence of context}
\label{subsec:context}
As an extension to the previous subsection, we present the confusion caused by visual patterns from the context. Figure~\ref{fig:context} shows the images and corresponding ground truth labels on the left. On the right, the representative image for the predicted \posts{} corresponding to them are shown. In the first case, presence of \emph{sky} region, landscapes makes the model to consider the two classes to be visually closer. In case of other samples of \emph{Yurt} category, we observe that the shape of the \emph{Yurt} being very similar to that of the \emph{Alp}. The visual similarity given by the model for the two classes (sec.~\ref{subsec:qivis}) is $0.137$. Similarly, in the second example, presence of the bowl(s) in the \emph{eggnog} sample makes the model to confuse between the two classes. Almost all the samples from \emph{eggnog} class have a very close context as that of \emph{mixing-bowl}. In this case, the visual similarity is $0.25$.
\subsection{Confusion among the fine-grained categories}
\label{subsec:dog-breeds}
In this subsection, we analyse the confusion of a CNN model among the fine-grained visual categories under various adversarial attacks. In particular, we chose the $117$ dog breed categories and GoogLeNet for this analysis. We consider $800$ validation samples belonging to these categories and compute the percentage of intra-dog category confusions. Figure~\ref{fig:dog-breeds}
shows the $\%$ of samples that are confused among these fine-grained categories, i.e., foolings in which a dog sample is misclassified as another dog category. Note that the confusion caused within the fine-grained categories by the image specific attacks is significantly high. Also, in spite of the existence of `dominant labels', image agnostic attacks also fool the CNN among the fine-grained categories. %This behaviour sort of suggests that the current attacks are predominantly `visual' in nature.  Note that our analysis includes some of the strong and state-of-the-art attacks such as PGD~\cite{madry-iclr-2018}, CW~\cite{cw-sp-2017}, NAG~\cite{mopuri-cvpr-2018}.
%-------------------------------------------------------------------------
\section{Conclusion}
\label{sec:conclu}
%Vulnerability of the AI systems to the adversarial attacks is not acceptable. It is essential to analyse the trained models and existing attacks to throw more light on the process and build robust models. In this paper, we discussed the shortcomings of the existing evaluation metric, the `fooling rate' to bring out the important aspects of the adversarial fooling. We presented novel metrics to measure the confusion caused by the attacks on both semantic and visual scales. This can help while deploying these models in real world environments with varying levels of tolerance to confusion. We have presented empirical evaluation of a range of adversarial attacks that include all varieties such as image specific, agnostic, iterative, and targeted. Also, we presented the analysis of CNN models from all the different families of architecture. Our metrics help to bring out the subtle differences to understand how different adversarial attacks realize the fooling.
In this paper, we challenge the current consensus in the field of using \emph{fooling rate} alone as a metric for evaluating the quality of an adversarial attack. We introduce three additional metrics $FR@K$, $QI\text{-}Wup$, and $QI\text{-}Vis$ that capture three different aspects of the fooling. %Using these, we have presented an empirical evaluation of a range of adversarial attacks that include varieties such as image-specific, agnostic, iterative, etc. 
They helped in bringing out previously unknown strengths and weaknesses of these attacks which would be helpful while deploying the CNN models in real-world environments. %These can also act as tools to quickly understand how different adversarial attacks realize the fooling. %We believe that the proposed metrics and the experimental analysis clearly support the objective of the draft which is \textit{to demonstrate the shortcoming of `fooling rate'} and propose new ones \textit{to fill the gap}.Note that unlike existing works, our work is the first to advocate the necessity for better metrics (Fig.~\ref{fig:fooling-rate-shortcomings}), and introduced multiple such metrics.
To the best of our knowledge, none of the existing works evaluates with metrics other than `fooling rate'. %As revealed by the proposed metrics, current attacks are designed solely to achieve high `fooling rate' and fail to realize fooling in the true sense.
We list some of the important inferences drawn from our work:

 %We were able to draw the following conclusions based on the analysis of CNN models from all the different families of architecture.
\begin{itemize}[noitemsep]
    %\item %In case of image specific attacks, they often confuse among the `visually' similar classes that are ranked high before the attack. However, the surprisingly high semantic confusion can be a result of the similar visual patterns (e.g. \emph{porcupine}-\emph{hay} as in Figure~\ref{fig:vis-sem}) that exist across different categories.
    %The image specific attacks, often confuse among the ‘visually’ similar classes (sec.~\ref{subsec:dog-breeds}) that are ranked high before the attack. However, in some cases, surprisingly high semantic confusion can be a result of the similar visual patterns (e.g. tennis ball - granny smith as in Figure~\ref{fig:vis-sem}) that exist across different categories. Nevertheless, we also notice remarkable differences in the `semantic' confusion caused by them.
    %\item In case of image-agnostic attacks, the existence of dominant labels causes significant `visual' and `semantic' confusion. However, the success rate of these attacks is relatively less compared to the image specific counter parts. Also, the relatively high semantic gap between the \emph{pre} and \posts{}, and limited set of \posts{} might alarm the users about an attack.
    %\item  
    \item  Our experimental results bring out the usefulness of the new metrics by clearly differentiating attack behaviours. For instance, AuC computed from the $FR@K$ graphs (Tab.~\ref{tab:auc}) reveal that the attacks vary in their ability to reduce the confidence assigned to the \pre{}. PGD demonstrates significantly higher AuC values suggesting its superiority over the others. This is interesting and in line with the observation that PGD attack is the most robust against current adversarial defenses~\cite{madry-iclr-2018}, which might be attributed to its ability to reduce the confidence to the \pre{}. Similarly, the observation that Deep Fool (and CW) attack generally swaps the top $2$ labels is non-trivial with only `fooling rate'. % Hence our metrics provide new tools which can clearly distinguish attack strengths when various important attacks behave similarly with respect to `fooling rate'. 
    \item \emph{QI-Wup} metric (Fig.~\ref{fig:qiwup-expts}) reveals that some of the attacks (e.g. FGSM based) that are less effective with respect to `fooling rate', are comparatively more severe on a semantic scale than their counter parts. On the other hand, attacks such as Deep Fool while achieving a high `fooling rate' inflict least amount of semantic damage. Clearly, when analysing the semantic damage inflicted by the attacks (or incurred by models), fooling rate \emph{can not} serve the need.
    \item Our experiments reveal (Sec.~\ref{subsec:dog-breeds}, Fig.~\ref{fig:qiwup-expts}) that some of the strongest adversarial attacks such as PGD achieve significant fooling $(>85\%)$ via confusing the models among visually similar, fine-grained categories which are only $\sim12\%$ of the total categories. Without this information, higher fooling rates achieved by these attacks may project the classifiers as unsophisticated to the community and more importantly to the policy makers. Similarly, attacks such as CW and DeepFool, in spite of resulting a very high (top-1) fooling rate, cause significantly lesser visual and semantic confusion. On the other hand, relatively simple attacks such as FGSM and FGSM-LL cause higher visual and semantic damage. These aspects throw new light on the attacks and the way they achieve the fooling. 
    %\item Our discussion section (sec.~\ref{sec:disc}) draws evidence to argue that the current adversarial attacks confuse the models among `visually' similar categories more frequently than not. Further, in cases where they achieve fooling across visually dissimilar categories (i.e., higher semantic damage), we found that the similar looking visual patterns and context might be responsible.
    %\item We experimented with one of the largest (public) computer vision datasets, multiple popular architectures, and a variety of important adversarial attacks making it a strong component of the draft.
    %\item We can observe that most of the existing attacks are predominantly `visual' in nature, i.e., the attacks are not designed to cause a large semantic confusion (except in the cases of targeted attacks), rather, they exploit the visual similarities among different categories to realize the \emph{fooling}. Nevertheless, we also notice remarkable differences in the `semantic' confusion caused by them.
\end{itemize} 

%-------------------------------------------------------------------------
{\small
\bibliographystyle{ieee}
\bibliography{mybibliography}
}

%%%%%% Appendix  %%%%%
\section{Supplementary}
\subsection{Extension To Targeted Attacks}
We believe that the proposed metrics can easily be extended to targeted attacks. A targeted attack is considered to be successful when it fools a model into falsely predicting a specific target label for the adversarial image and is considered to be a failure otherwise.  However the attack can be considered a success even if it fools a model into falsely predicting a label that is semantically/visually similar to the target class.  Hence, the proposed metrics in eq.3, and 5 can be modified for targeted attacks as follows: 
 
 {\scriptsize
\begin{equation*}
    QI\text{-}Wup\text{-}Targeted=\begin{cases}
    1, & \text{if $Wup(\text{\tar{}},\text{\post{}})<T_s$}.\\
    0, & \text{otherwise}.
  \end{cases}
\end{equation*}
}
\vspace{-0.25cm}
{\scriptsize
\begin{equation*}
    QI\text{-}Vis\text{-}Targeted=\begin{cases}
    1, & \text{if $Vis(\text{\tar{}},\text{\post{}})<T_v$}.\\
    0, & \text{otherwise}.
  \end{cases}
\end{equation*}
}

\subsection{Visual similarities learned by the CNNs}
\begin{figure}[h]
    \centering
    \includegraphics[width=0.45\textwidth]{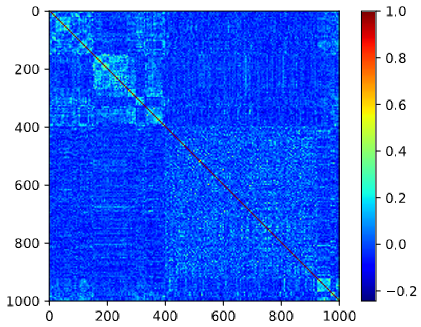}
    %googlenet-vis-similarities.pdf
    \caption{Pairwise visual similarities learned by by GoogLeNet for the $1000$ categories of ILSVRC dataset.}
    \label{fig:googlenet-vis-similarities}
\end{figure}

\begin{figure}[]
    \centering
    \includegraphics[width=0.45\textwidth]{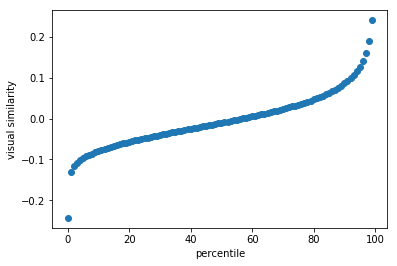}
    \caption{Percentile graph for the visual similarities computed for GoogLeNet trained. X-axis shows the $\%$ of visual similarities that are less than a chosen value on Y-axis.}
    \label{fig:vis-knee-point}
\end{figure}

In this section we present more information about the visual similarities introduced in sec $2$ of the main draft. We consider GoogLeNet for the analysis presented here. Note that the visual similarities are model specific and might vary across different models. Figure~\ref{fig:googlenet-vis-similarities} shows the pairwise visual similarities learned by the GoogLeNet. Figure~\ref{fig:vis-knee-point} shows the percentile graph for the visual similarities computed among the $1000$ object classes of the ILSVRC dataset. The graph shows percentage of the pairwise visual similarities (on x-axis) that have a value less than a given similarity on y-axis. The minimum visual similarity is around -$0.25$ and the maximum is $1.0$. Note that for better visualization, we present a zoomed-in plot. The threshold $(T_v)$ for computing the mean visual confusion (sec. $3$ of the main draft) was chosen to be $0.1$ since $95\%$ of the categories have visual similarities less than $0.1$ as per the percentile graph. We can observe a ``knee point'' at the the $95\%$ percentile on the x axis.
% ######           Fig-1 (googlenet vis similarities) begin       ####################
% ######           Fig-1 (googlenet vis similarities) end       ####################
\subsection{Semantic similarities across the ILSVRC categories}
\begin{figure}[]
    \centering
    \includegraphics[width=0.45\textwidth]{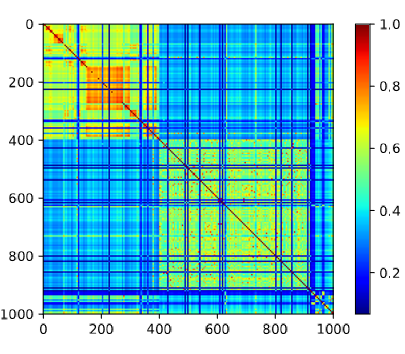}
    %wup-similarity.pdf
    \caption{Pairwise Wup (Wu-Palmer) similarities computed for the $1000$ object categories present in ILSVRC dataset.}
    \label{fig:wup-similarities}
\end{figure}

\begin{figure}[]
    \centering
    \includegraphics[width=0.475\textwidth]{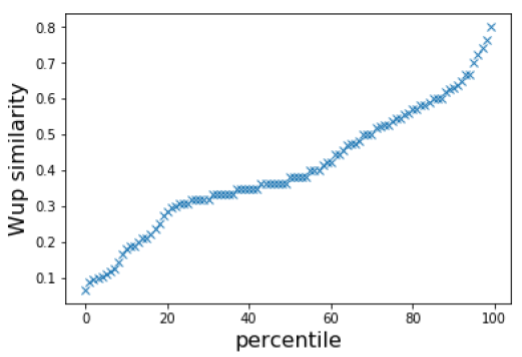}
    \caption{Percentile graph for the semantic (Wu-Palmer) similarities computed for the object categories present in ILSVRC dataset. X-axis shows the $\%$ of semantic similarities that are smaller than a chosen value on Y-axis.}
    \label{fig:sem-knee-point}
\end{figure}

In this section we present the semantic similarities among the $1000$ object categories present in the ILSVRC dataset. We have used Wu-Palmer (Wup) similarity measure for computing the semantic similarity. Note that these similarities are independent of the CNN model. Figure~\ref{fig:wup-similarities} shows the pairwise Wup similarity computed for all the $1000$ object categories in ILSVRC dataset. We can observe high similarity values among the dog breed categories ranging between the class indices $150$ and $270$ (the red square block). Note that the CNN models also learn high visual similarities among these fine-grained categories. We can observe similar block in Figure~\ref{fig:googlenet-vis-similarities}. Figure~\ref{fig:sem-knee-point} shows the percentile graph for the computed semantic similarities. It is observed that the Wup similarities are in general high. For instance, $50\%$ of the similarities are higher than $0.4$. Therefore, the semantic similarities are in general large compared to their visual counter parts.  Hence, the threshold $(T_s)$ chosen to compute the mean semantic confusion (sec. $3$ of the main draft) is $0.7$. Note that this is the knee point equivalent the one chosen for visual similarities, i.e., $95\%$ of the categories have semantic similarities less than $0.7$.
% ######           Fig-2 (Knee-point) begin       ####################
% ######           Fig-2 (knee-point) end       ####################
% ######           Fig-3 (wup similarities) begin       ####################
% ######           Fig-3 (wup similarities) end       ####################
% ######           Fig-4 (sem Knee-point) begin       ####################
% ######           Fig-4 (sem Knee-point) end       ####################
\subsection{Ablations for CW attack}
\label{sec:conclu}
\begin{figure}[]
    \centering
    \includegraphics[width=0.45\textwidth]{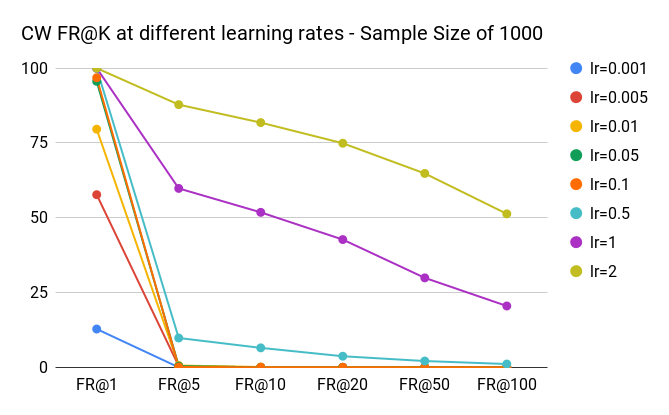}
    \caption{Effect of learning rate in CW on our FR@K metric.}
    \label{fig:fr-vs-lr-cw}
\end{figure}

\begin{figure}[]
    \centering
    \includegraphics[width=0.45\textwidth]{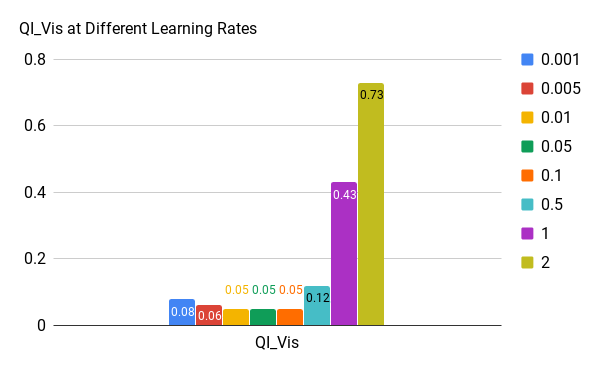}
    \caption{Effect of learning rate in CW on our QI-Vis metric.}
    \label{fig:qivis-vs-lr-cw}
\end{figure}

\begin{figure}[]
    \centering
    \includegraphics[width=0.45\textwidth]{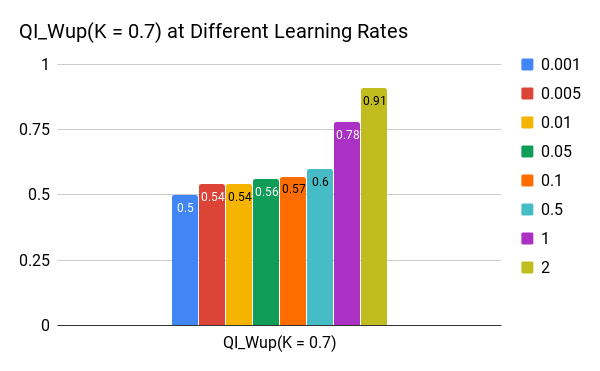}
    \caption{Effect of learning rate in CW on our QI-Wup metric.}
    \label{fig:qiwup-vs-lr-cw}
\end{figure}

The choice of learning rate (for optimizing the objective function) has not been discussed in case of Carlini-Wagner attack in its original paper~\cite{cw-sp-2017}. We noticed significant variance in the quality of the attack at different learning rates. Although there was an increase in the performance in terms of all three metrics as we increased the learning rate, we noticed large distortion in the image quality beyond learning rate of $0.1$ and hence a learning rate of $0.1$ was chosen. The performance metrics for GoogLeNet architecture for all the $3$ proposed metrics are shown in Figures~\ref{fig:fr-vs-lr-cw}, \ref{fig:qivis-vs-lr-cw} and~\ref{fig:qiwup-vs-lr-cw}. To summarise, we observed large visual contamination (and $l_p$ norm of the contamination of the crafted perturbation) for learning rates higher than $0.1$. Therefore, in order to provide a reasonable comparison with other attacks, we report (in the main draft) the metrics with the default values of the hyper-parameters. However, please note that the attack's performance can strongly vary against the strength of the perturbation which is not directly limited by their objective.

\end{document}